%% file: root.tex
\documentclass[letterpaper, 10 pt, conference]{ieeeconf}  % Comment this line out if you need a4paper

\IEEEoverridecommandlockouts                              % This command is only needed if 
\usepackage{amsmath} % assumes amsmath package installed
\usepackage{amssymb}  % assumes amsmath package installed
\usepackage{graphicx}
\usepackage[font=footnotesize]{caption}
\usepackage[font=footnotesize]{subcaption}
\usepackage{url}
\usepackage{tabularx}
\usepackage{booktabs}
\usepackage{mathtools}
\usepackage[sort,compress]{cite}
\usepackage[colorinlistoftodos,prependcaption,textsize=tiny]{todonotes}
\usepackage{flushend}

\include{tikz_setup}

\usepackage{url}

\newboolean{arxiv}
\setboolean{arxiv}{true}

\def\doi{DOI: \href{https://doi.org/XX.YYYYY/X.YYYYY}{XX.YYYYY/X.YYYYY}}
\newcommand\copyrighttext{%
 \footnotesize \textcopyright \the\year{} IEEE. Personal use of this material is permitted.
  Permission from IEEE must be obtained for all other uses, in any current or future   media, including reprinting/republishing this material for advertising or promotional   purposes, creating new collective works, for resale or redistribution to servers or   lists, or reuse of any copyrighted component of this work in other works.
  }
\newcommand\copyrightnotice{%
\begin{tikzpicture}[remember picture,overlay]
\node[anchor=south,yshift=10pt] at (current page.south) {\fbox{\parbox{\dimexpr\textwidth-\fboxsep-\fboxrule\relax}{\copyrighttext}}};
\end{tikzpicture}%
}

\title{\LARGE \bf
DeepSeeColor: Realtime Adaptive Color Correction for Autonomous Underwater Vehicles via Deep Learning Methods
}

\author{Stewart Jamieson$^{1,2}$, Jonathan P. How$^{2}$, and Yogesh Girdhar$^{3}$% <-this % stops a space
\thanks{*This work was supported by NSF-NRI Award Number 1734400, and by grants from NVIDIA. This work utilized an NVIDIA RTX A6000 GPU.}% <-this % stops a space
\thanks{$^{1}$S. Jamieson is with the MIT-WHOI Joint Program in Oceanography/Applied Ocean Science and Engineering
        {\tt\small sjamieson@whoi.edu}}%
\thanks{$^{2}$S. Jamieson and J. P. How are with the Department of Aeronautics and Astronautics at the Massachusetts Institute of Technology (MIT)
        {\tt\small \{sjamieson, jhow\}@mit.edu}}%
\thanks{$^{3}$Y. Girdhar is with the Woods Hole Oceanographic Institution (WHOI) Applied Ocean Physics and Engineering Department
        {\tt\small yogi@whoi.edu}}%
}

\begin{document}

\maketitle
\thispagestyle{empty}
\pagestyle{empty}

%%%%%%%%%%%%%%%%%%%%%%%%%%%%%%%%%%%%%%%%%%%%%%%%%%%%%%%%%%%%%%%%%%%%%%%%%%%%%%%%
\begin{abstract}
\ifthenelse{\boolean{arxiv}}{\copyrightnotice}{}%
Successful applications of complex vision-based behaviours \textit{underwater} have lagged behind progress in terrestrial and aerial domains. This is largely due to the degraded image quality resulting from the physical phenomena involved in underwater image formation. \textit{Spectrally-selective light attenuation} drains some colors from underwater images while \textit{backscattering} adds others, making it challenging to perform vision-based tasks underwater.  State-of-the-art methods for underwater color correction optimize the parameters of image formation models to restore the full spectrum of color to underwater imagery.  However, these methods have high computational complexity that is unfavourable for realtime use by autonomous underwater vehicles (AUVs), as a result of having been primarily designed for offline color correction.  Here, we present \textit{DeepSeeColor}, a novel algorithm that combines a state-of-the-art underwater image formation model with the computational efficiency of deep learning frameworks.  In our experiments, we show that DeepSeeColor offers comparable performance to the popular ``Sea-Thru'' algorithm~\cite{Akkaynak2019} while being able to rapidly process images at up to 60Hz, thus making it suitable for use onboard AUVs as a preprocessing step to enable more robust vision-based behaviours.
\end{abstract}

\section*{Open-Source Software}

% Publication of this work will be accompanied by a public release of the \textit{DeepSeeColor} source code under a permissive license, with encouragement for community collaboration on further development. It will also be released with a dataset containing high-resolution stereo imagery of coral reefs in the US Virgin Islands, to enable broader efforts in this and related research directions.
The datasets collected for and used in this paper are available along with an implementation of DeepSeeColor at:\\
\url{https://warp.whoi.edu/deepseecolor/}

%%%%%%%%%%%%%%%%%%%%%%%%%%%%%%%%%%%%%%%%%%%%%%%%%%%%%%%%%%%%%%%%%%%%%%%%%%%%%%%%
\section{Introduction}

The development of autonomous underwater vehicles (AUVs) has trended towards more complex and adaptive vision-based behaviours, enabled by the increasing performance and energy efficiency of onboard micro-computing resources over time.
Many such behaviours, including visual target tracking~\cite{Cai2023, Yoerger2021,Katija2021,Hao2022}, adaptive exploration of benthic environments~\cite{Jamieson2020,Guerrero2021,Girdhar2016a,Bonin-Font2015,Joshi2022,Manjanna2016,Rao2016}, autonomous docking~\cite{Singh2020,Matsuda2019,Yazdani2020}, and diver following~\cite{Chou2020} and assistance~\cite{Stilinovic2015,GomezChavez2019,Pelletier2022,Birk2022}, strongly depend on an AUV's capability for visual perception and understanding.
Unfortunately, underwater image quality is generally lower and more variable than the quality of similar images taken in air, making vision-based autonomous behaviours far more challenging to perform underwater than in other domains.

\begin{figure}
    \centering
    \includegraphics[width=\columnwidth]{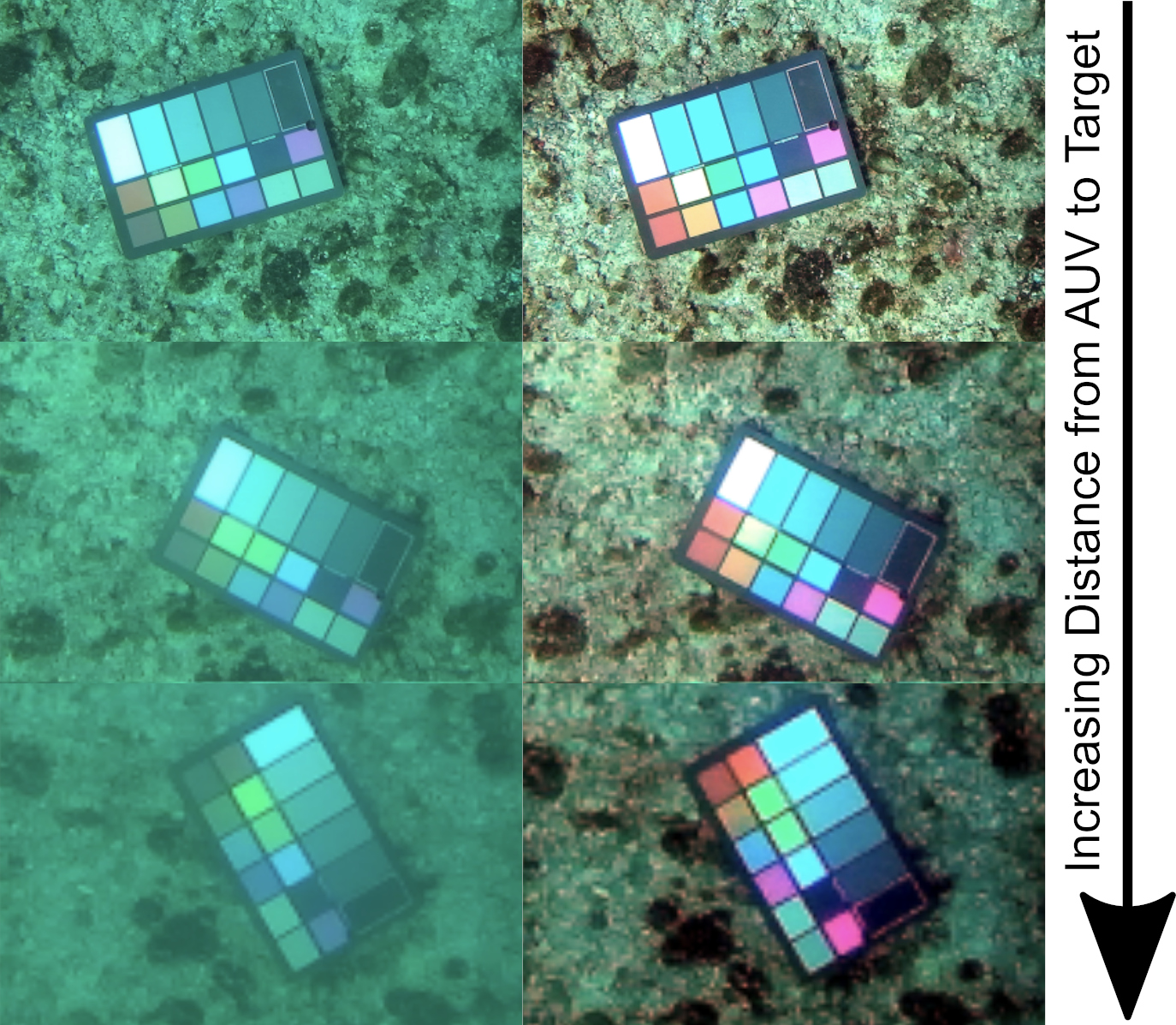}
    \caption{Best viewed on a screen, in color. Images taken in underwater environments (left column) suffer from severe loss of colors due to the combination of spectrally-selective light attenuation and backscattering due to particles in water. This effect is more pronounced as the distance between the camera and imaging target increases. This paper presents DeepSeeColor, an efficient technique to reconstruct true-color images (right column) for use in realtime onboard decision making by an autonomous underwater vehicle.}
    \label{fig:back_atten}
\end{figure}

The most prominent image quality issues result from spectrally-selective light attenuation and backscattering. These phenomena cause images taken underwater to be lacking in certain colors, typically red, and to have an excess of others, such as blue and green, as seen in Fig.~\ref{fig:back_atten}. The magnitude of these effects is correlated with the camera's water depth and range to the target; objects in the image that are shallow and close to the camera will have more of their natural color than objects that are deeper or further away. However, the specific color distortions in an image depend also on the presence of other light sources (e.g., the sun), and on the optical properties of the camera and the body of water in which the image was captured. 

The performance of computer vision algorithms for image classification and 3D reconstruction is severely degraded by color distortions, regardless of whether those algorithms are based on deep neural networks or on more traditional feature detection and matching methods~\cite{Zhou2014,De2021}. Even learning-based systems trained to be robust to targets within some range of distances, lighting conditions, and water masses will generally not be robust to real-world targets outside of the training distribution. This makes realtime, adaptive, \textit{in situ} color correction a necessary prerequisite for successful vision-based AUV tasks and behaviours, especially in potentially safety-critical applications such as diver assistance. This is a gap in the underwater vision field; while previous works have focused on high-quality offline color reconstruction~\cite{Carlevaris-Bianco2010,Yang2011,Chiang2012,Berman2017,Bryson2016,Akkaynak2019}, to our knowledge no methods have been developed specifically for realtime applications.

This paper presents \textit{DeepSeeColor}, a novel method for realtime and adaptive color correction of images onboard an AUV. This realtime preprocessing enables more robust execution of \textit{in situ} autonomous behaviours such as object recognition and tracking, semantic mapping, and mission summarization. \textit{DeepSeeColor} significantly improves upon the successful ``Sea-Thru'' algorithm~\cite{Akkaynak2019} in terms of computational efficiency. It achieves this improvement through using highly efficient gradient-based optimization methods to learn the weights of two simple convolutional neural networks, where these weights correspond to a physics-based image formation model's unknown parameters. This novel approach replaces far more computationally expensive image processing operations, while leveraging GPU acceleration and other built-in performance optimizations provided by popular deep learning frameworks. We further present validation of DeepSeeColor on the dataset provided with~\cite{Akkaynak2019}, as well as on a new dataset of stereo-imagery collected from AUV deployments at coral reef sites in the US Virgin Islands.

\section{Background \& Related Works}

\subsection{Underwater Image Formation}

A digital camera system creates images by measuring the intensity of incoming light reflected from various targets in its lens' field of view. Assuming the light originates from a broadband source similar to the sun, the wavelengths of the reflected light accurately represent the colors of the targets. A color filter array on the camera's image sensor enables measuring the intensity of light within the red, green, and blue wavelength ``channels''\footnote{These channels are roughly defined as the wavelength ranges 575-725nm (red), 475-625nm (green), and 400-550nm (blue), respectively.} at each pixel.

Regardless of wavelength, the apparent intensity of light decreases in relation to an observer's distance from its source according to the inverse-square law.
However, the presence of an intervening medium further attenuates the apparent intensity of a target through \textit{absorption} and \textit{scattering}.
Furthermore, many media are \textit{spectrally-selective}: they preferentially attenuate some wavelengths of light over others.
Water is much more spectrally-selective than air: red light is attenuated over a distance of 1m about as much as green light is over 10m, or blue light is over 100m~\cite{Pegau1997}. % figure?

\textit{Backscattering} is the process by which particles in the water column between the target and camera reflect light from sources other than the target into the camera. This is particularly problematic at shallow depths ($<$100m), long target ranges ($>$1m), and in turbid waters.\footnote{Note ``depth'' refers to the vertical distance of the target from the water surface, while ``range'' refers to the distance from the camera to the target.} At shallow depths, light from the surface (e.g., sunlight) is scattered in all directions by tiny particulate in the water; the spectrum of this scattered light consist of wavelengths that were not absorbed higher up in the water column, and thus tends to be blue or blue-green. The greater the range to the target, the more intervening particulate there is to backscatter this light into the camera; the resulting ``haze'' in the image makes it more difficult to see the target, especially when it saturates the camera's dynamic range.

We denote the intensity of light in channel $c\in\{\texttt{R}, \texttt{G}, \texttt{B}\}$ emitted by the target along the ray to pixel $(i,j)$ in the camera image sensor as $J_c(i,j)$, and the intensity measured by the sensor at that pixel as $I_c(i,j)$.
% \footnote{Thus, for a camera with even spectral response imaging a target in a vacuum, we would have $I_c(i,j)=J_c(i,j), \forall c, i, j$.}
Spectrally-selective light attenuation and backscattering can be modelled together using the underwater image formation model,
\begin{equation}\label{eq:image-formation}
    I_c(i,j) = J_c(i,j) A_c(i,j) + B_c(i,j),
\end{equation}
where $A_c(i,j)$ and $B_c(i,j)$ represent wavelength-dependent light attenuation and backscatter, respectively. This reflects the form of models used by many works in underwater color reconstruction~\cite{Carlevaris-Bianco2010,Yang2011,Chiang2012,Berman2017,Bryson2016}, which take $A$ and $B$ to be
\begin{align}
    A_c(i, j) &= \exp\left(-a_c \cdot z_{i,j}\right),\label{eq:wrong-attenuation} \\
    B_c(i, j) &= \gamma_c^\infty \left(1 - A_c(i,j)\right),\label{eq:wrong-backscatter}
\end{align}
where $z_{i,j}$ is the range of the target. The values of $a_c, \gamma_c^\infty \in \mathbb{R}_{\ge0}$ are determined by the camera system and environmental parameters, including as the water type, target reflectance, illumination sources, image sensor characteristics, and camera depth, which are, for now, all assumed to be fixed.

More recently,% Akkaynak \& Treibitz
~\cite{Akkaynak2018}
found that Eq.~\eqref{eq:wrong-attenuation}, derived from an atmospheric dehazing model, neglects the range-dependence of underwater light attenuation coefficient $a_c$, and incorrectly assumes that the coefficients governing the range-dependence of attenuation and backscattering are the same. \cite{Akkaynak2018} presented a new model capable of capturing these complexities, 
\begin{align}
    A_c(i,j) &= \exp\left(-a_c(z_{i,j}) \cdot z_{i,j}\right),\label{eq:sea-thru-attenuation} \\
    B_c(i,j) &= \gamma_c^\infty \left(1 - \exp\left(-\beta_c  z_{i,j}\right)\right),\label{eq:sea-thru-backscatter}
\end{align}
with scalars $\gamma_c^\infty, \beta_c \in \mathbb{R}_{\ge0}$ and a parametric function $a_c(z)$.\footnote{$\beta_c$ is most accurately modelled as a function of the range $z$, like $a_c(z)$, but this effect is negligible within a fixed water type~\cite{Akkaynak2018}.}$^,$\footnote{Details on the interpretation of each parameter are given in~\cite{Akkaynak2019}.}
This is the model used by Sea-Thru~\cite{Akkaynak2019} and DeepSeeColor, and will be explored further in Section~\ref{sec:methods}.

\begin{figure}
    \centering
    \includegraphics[width=\columnwidth]{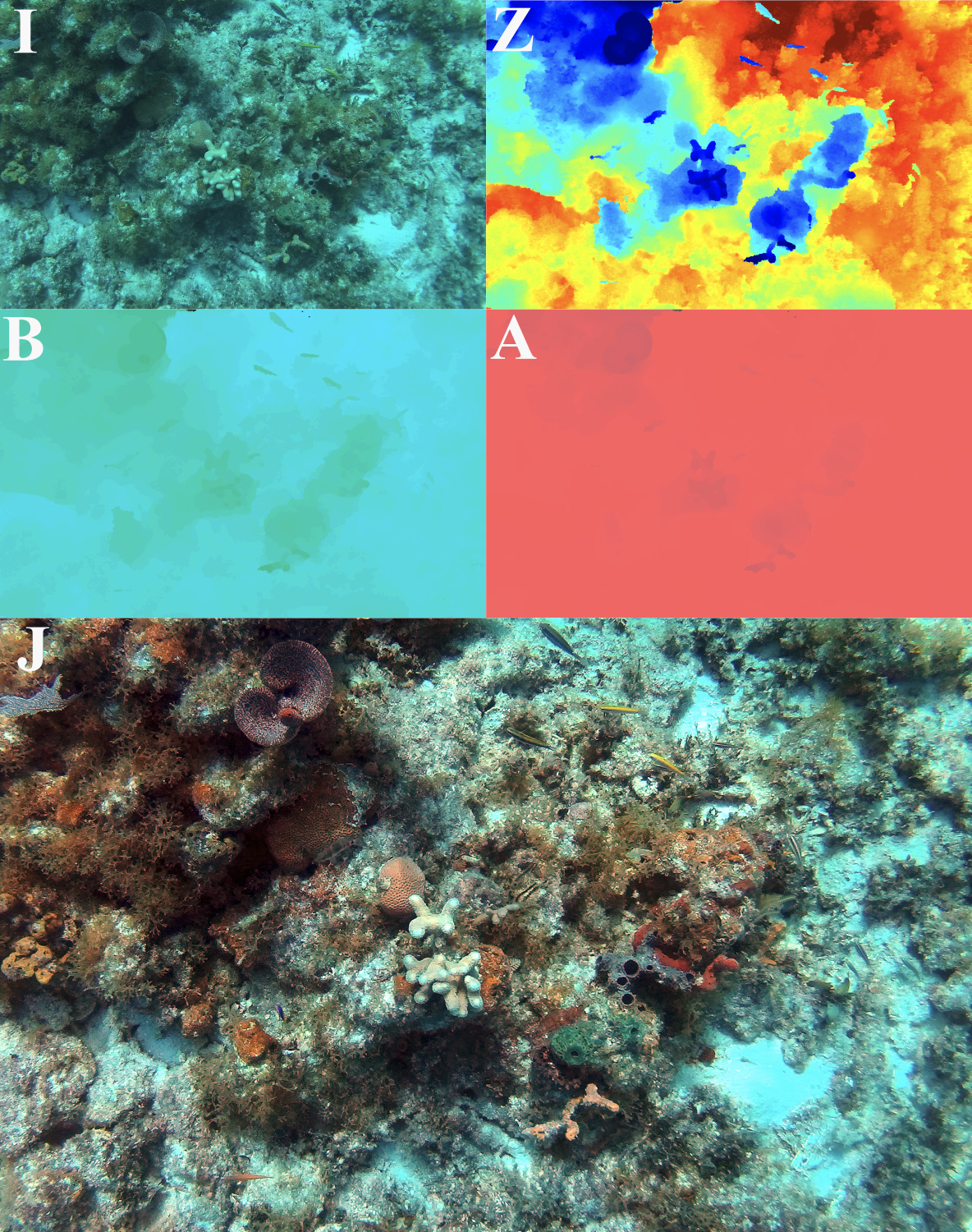}
    \caption{Examples of an input image $I$ and corresponding range map $Z$ (color-coded such that blue is ``close'' and red is ``far''), best viewed on a screen in color. The DeepSeeColor method efficiently learns to approximate the backscatter image $B$ and attenuation factor image $A$ (both shown with contrast enhanced), enabling recovery of the true-color image $J$.}
    \label{fig:model}
\end{figure}

\subsection{Methods for Color Reconstruction}

Spectrally-selective light attenuation and backscattering have been considered by underwater color reconstruction algorithms since the seminal works of \cite{Carlevaris-Bianco2010,Yang2011,Chiang2012}. %Carlevaris-Biano et al.\cite{Carlevaris-Bianco2010}, Yang et al.~\cite{Yang2011}, and Chiang \& Chen~\cite{Chiang2012}.
These works presented each presented a method to estimate their respective image formation model parameters from a single image. Given these parameter estimates, the true color of the target at pixel location $(i,j)$ can be reconstructed as
\begin{equation}
    J_c(i,j) = D_c(i,j) A_c(i,j)^{-1},
\end{equation}
where $D_c(i,j) \coloneqq I_c(i,j) - B_c(i,j)$ denotes the ``direct'' signal. The components of this approach to color reconstruction are depicted in Fig.~\ref{fig:model}. 
More recent works have explored techniques for better estimating the parameters $a_c,\gamma_c^\infty$ through leveraging additional information, such as the optical properties of known water profiles~\cite{Berman2019}.

Among the most relevant of prior works is% Bryson et al.
~\cite{Bryson2016}, which recognized that leveraging an accurate range map, which describes the distance of each pixel in an image, could better constrain estimates of the image formation model parameters.
% \footnote{Previous works had looked at the inverse problem of using attenuation and backscatter estimates to predict the range of each pixel~\cite{Carlevaris-Bianco2010,Chiang2012}.}
The method was designed for use on AUV collected data, and applied structure-from-motion estimation to generate range maps based on the overlapping field-of-view and real-world camera displacement between pairs of captured images~\cite{Bryson2016}. This displacement was estimated using navigational information collected by other sensors on the AUV, but estimation could be avoided by using a synchronized stereo-camera system to simultaneously capture pairs of images with a known, fixed camera displacement.

The previously discussed methods were developed for estimating the parameters of the traditional underwater image formation model described by Eqs.~\eqref{eq:wrong-attenuation} and~\eqref{eq:wrong-backscatter}. Estimating the parameters of the improved underwater image formation model presented in~Eqs.~\eqref{eq:sea-thru-attenuation} and~\eqref{eq:sea-thru-backscatter} originally required the use of multiple color chart calibration targets \textit{in situ}~\cite{Akkaynak2018}. The idea of using range maps, instead of color charts, to guide parameter estimation was adapted by the \textit{Sea-Thru} algorithm~\cite{Akkaynak2019}, which enabled learning the improved model parameters using only the captured image and range map.

There has also been progress in learning-based methods for color correction~\cite{Porav2018}, including for underwater imagery~\cite{Li2017,Fabbri2018}. However, these methods do not guarantee consistency or accuracy in the color reconstruction of an image stream, or images captured in different environmental conditions. This makes them less suitable than physics-based color correction methods for use in supporting realtime, possibly safety-critical, autonomous underwater behaviours.

\subsection{AUV Vision System Considerations}

Though~\cite{Bryson2016} specifically explored color reconstruction for images \textit{collected} by AUVs, to our knowledge no prior works have developed robust, physics-based underwater color reconstruction methods intended to run \textit{onboard} an AUV's highly constrained computational resources in realtime.
In fact, each of the aforementioned color reconstruction methods relies on solving optimization problems with computational complexity that is at least linear in the number of pixels in the input image~\cite{Carlevaris-Bianco2010,Yang2011,Chiang2012,Berman2017,Bryson2016,Akkaynak2019}. The significant computaional complexity of Sea-Thru~\cite{Akkaynak2019}, the only prior work to solve the parameters of the improved image formation model in Eqs.~\eqref{eq:sea-thru-backscatter} and~\eqref{eq:sea-thru-atten-coeff}, will be explored further in Section~\ref{sec:methods}.
% \footnote{For example, at the time of writing, the only public implementation of the Sea-Thru algorithm \label{ft:sea-thru}\url{https://github.com/hainh/sea-thru} requires minutes to process a single FHD (1920x1080) color image at full resolution on a desktop workstation.}

Computationally expensive methods could be acceptable if the image formation parameters only needed to be computed once. However, prior works have found that these image formation parameters can change significantly in time and space due to factors including variation in depth, lighting conditions, exposure time, turbidity, and imaging angle~\cite{Akkaynak2018,Akkaynak2019}. This demands the usage of an \textit{adaptive} color correction method that is robust to changes in lighting and other environmental parameters, and can be run in realtime on each image as it is collected.

\section{The DeepSeeColor Method}\label{sec:methods}

\begin{figure}
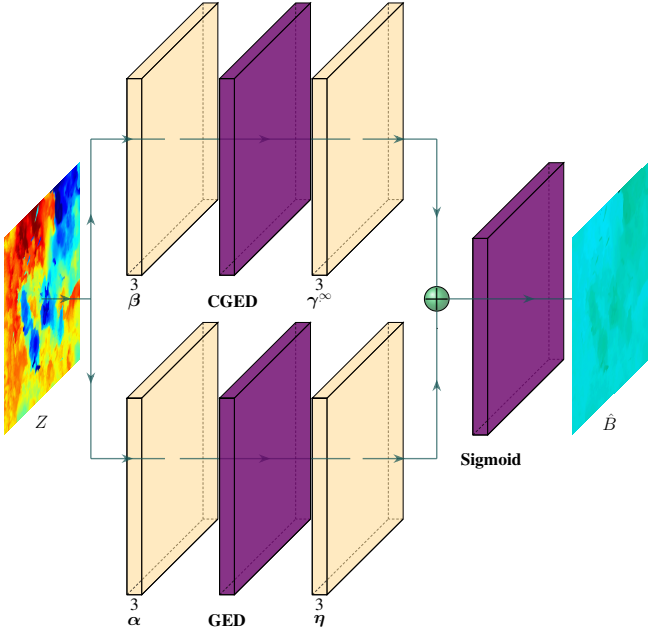

    \centering
    \include{figs/backscatter_net}
    \caption{The backscatter net inputs a range map $Z$ (visualized in color) and estimates the corresponding backscatter image $\hat{B}$. The kernel parameters in each convolutional layer map to the parameters of the backscatter estimation model in Eq.~\eqref{eq:sea-thru-backscatter-residual}.}
    \label{fig:backscatter-net}
\end{figure}

The DeepSeeColor method estimates the backscatter and attenuation parameters of the underwater image formation model from~\cite{Akkaynak2018} using a sequence of two convolutional neural networks~\cite{LeCun2015} depicted in Figures~\ref{fig:backscatter-net} and~\ref{fig:attenuation-net}, respectively. The networks are trained under self supervision using the captured image $I$ and range map $Z$; this training process can take advantage of highly energy-efficient deep learning hardware accelerators increasingly found onboard autonomous platforms~\cite{Reuther2020}.

% \subsection{The Gated Exponential Decay Activation Function}

%  These functions enable modelling the underwater image formation model from~\cite{Akkaynak2018} as a feed-forward neural network. %, but are not intended for general usage beyond this particular application.

% Like the popular ReLU activation function $\text{ReLU}(x) = \max\{0,x\}$, it is continuous but not differentiable at $x=0$.

\subsection{Backscatter Estimation}

% The goal of backscatter estimation is to estimate the (attenuated) ``direct signal'',
% \begin{equation}
%     \hat{D}_c(i,j) = I_c(i,j) - \hat{B}_c(z_{i,j})
% \end{equation}
% using the backscatter $\hat{B}_c(z_{i,j})$ estimated for a target seen in the pixel at location $(i,j)$ and with known range $z_{i,j}$. 

DeepSeeColor makes use of the same backscatter model as Sea-Thru,
\begin{equation}\label{eq:sea-thru-backscatter-residual}
    \hat{B}_c(i,j) = \gamma^{\infty}_c \left(1 - \exp\left(-\beta_c \cdot z_{i,j}\right) \right) + \eta_c \exp\left(-\alpha_c \cdot z_{i,j}\right),
\end{equation}
which corresponds to the backscatter model presented in Eq.~\eqref{eq:sea-thru-backscatter} augmented with a residual term, described in~\cite{Akkaynak2019}, characterized by two new scalar parameters $\eta_c,\alpha_c \in \mathbb{R}_{\ge0}$.
% \footnote{\cite{Akkaynak2019} provides some insights on when this residual term can be neglected.}
% $\hat{B}_c$ and $\alpha_c$ are notable as initial estimates of $B_c$ and the average $\alpha_c$ over the scene, respectively.
% The 12 unknown backscatter coefficients (4 per color channel) are estimated in Sea-Thru by binning the pixels in the image into 10 evenly spaced range windows, and assigning the darkest 1\% of pixels in each bin to the set $\Omega$. Assuming these pixels should all be true-black, the backscatter parameters are found by solving the optimization problem,
% \begin{equation}\label{eq:seathru-bs-opt}
% \begin{aligned}
% \min_{\boldsymbol{\gamma}^\infty,\boldsymbol{\beta},\boldsymbol{\eta},\boldsymbol{\alpha}} \quad & \sum_{(i,j)\in\Omega} \sum_{c} \left\vert I_c(i,j) - \hat{B}_c(i,j) \right\vert^2,\\
% \textrm{s.t.} \quad & \boldsymbol{\gamma}^\infty \in [0,1]^3, \boldsymbol{\beta} \in [0,5]^3, \\
%   & \boldsymbol{\eta} \in [0,1]^3, \boldsymbol{\alpha} \in [0,5]^3.\\
% \end{aligned}
% \end{equation}

\subsubsection{Inference}

DeepSeeColor performs backscatter estimation by using the neural network presented in Fig.~\ref{fig:backscatter-net}. 
We define a novel nonlinear activation function, Gated Exponential Decay (GED), as
\begin{equation}
    \text{GED}(x) \coloneqq \begin{cases}
        1 & x \le 0,\\
        \exp\left(-x\right) & x > 0.
    \end{cases}
\end{equation}
We also introduce its complement, $\text{CGED}(x) \coloneqq 1 - \text{GED}(x)$.
Observe that Eq.~\ref{eq:sea-thru-backscatter-residual} can now be rewritten as
\begin{equation}
    \hat{B}_c(Z) = \gamma^{\infty}_c \cdot \text{CGED}\left(\beta_c Z\right) + \eta_c \cdot \text{GED}\left(\alpha_c Z\right).
\end{equation}

With its final sigmoid activation layer removed, the network depicted in Fig.~\ref{fig:backscatter-net} computes $\hat{B}_c$ for the backscatter parameters encoded in its convolutional layers' kernels.
The operations $\beta_c Z$ and $\alpha_c Z$ for $c\in\{\texttt{r},\texttt{g},\texttt{b}\}$ are modelled as convolutions between the tensor $Z=\left[z_{i,j}\right]$ with dimensions $(W, H, 1)$ and a kernel of shape $(1, 1, 1, 3)$ to produce an output tensor with dimensions $(W, H, 3)$, where the values of the kernel elements correspond to $\boldsymbol{\beta}$ and $\boldsymbol{\alpha}$, respectively.\footnote{Note tensor shapes are given in order of (width, height, channels), while kernel shapes are given as (width, height, input channels, output channels).}
Similarly, multiplication by the coefficients in $\boldsymbol{\gamma}^\infty$ and $\boldsymbol{\eta}$ can be modelled as convolutions with kernels of dimension $(1,1,3,3)$, where each kernel is constrained to be diagonal and thus has only 3 free parameters, resulting in the scaling of each channel by the corresponding $\gamma^\infty_c$ or $\eta_c$. These parallel sequences of operations correspond to the top and bottom branches of the network in Fig.~\ref{fig:backscatter-net}, respectively.
The final sigmoid layer applies the activation function $\sigma(x)\coloneqq 1/\left(1 + e^{-x}\right)$, ensuring that the backscatter estimates remain bounded in $[0,1]$ regardless of any outliers in the range map.

\subsubsection{Training}

DeepSeeColor leverages the assumption from~\cite{Akkaynak2019,He2011} that in any given range interval there are some pixels in the image which should have zero intensity. It trains the backscatter network using a novel loss function,
\begin{equation}\label{eq:bs-loss}
\begin{aligned}
    \mathcal{L}_\text{bs}(\hat{D}) = \sum_{(i,j)} \sum_c &\big(\max\{\hat{D}_c(i,j), 0\} \\
    &\quad+ k \min\{\hat{D}_c(i,j), 0\}\big),
\end{aligned}
\end{equation}
with hyperparameter $k>1$ and $\hat{D}_c \coloneqq I_c - \sigma(\hat{B}_c)$.  %Computing the direct signal input $\hat{D}$ requires the color image $I$ corresponding to $Z$.

As $k\to\infty$, optimizing Eq.~\eqref{eq:bs-loss} becomes equivalent to finding backscatter parameters that minimize the minimum intensity of pixels at every range, while ensuring the network \textit{never} predicts that a pixel in the direct signal would have negative intensity. As $k\to1$, the loss function becomes tolerant of some pixels in the direct signal being predicted to have negative intensity, if that enables bringing more pixels closer to zero intensity. Intuitively, larger values of $k$ would produce more accurate backscatter estimates if the range map was noiseless and the assumption of zero-intensity pixels at every range was satisfied, but are also less robust to noise and outliers in the image and range map. We find empirically that a value of $k=1000$ produces accurate backscatter estimates.

For comparison, the backscatter parameters are estimated in Sea-Thru by binning the image pixels into 10 evenly spaced range windows, and assigning the darkest 1\% of pixels in each bin to the set $\Omega$~\cite{Akkaynak2019}. Assuming these pixels should have zero-intensity, the backscatter parameters are then found by solving the nonlinear optimization problem,
\begin{equation}\label{eq:seathru-bs-opt}
\begin{aligned}
\min_{\boldsymbol{\gamma}^\infty,\boldsymbol{\beta},\boldsymbol{\eta},\boldsymbol{\alpha}} \quad & \sum_{(i,j)\in\Omega} \sum_{c} \left\vert I_c(i,j) - \hat{B}_c(i,j) \right\vert^2.%,\\
% \textrm{s.t.} \quad & \boldsymbol{\gamma}^\infty \in [0,1]^3, \boldsymbol{\beta} \in [0,5]^3, \\
  % & \boldsymbol{\eta} \in [0,1]^3, \boldsymbol{\alpha} \in [0,5]^3.\\
\end{aligned}
\end{equation}
Constructing $\Omega$ using a bitonic merge sort has time complexity $O(NP^{-1}\log^2{N})$ for $N$ pixels in the input image, when run in parallel on a processor with $P \le N$ cores. In contrast, computing the loss function in Eq.~\eqref{eq:bs-loss} and backpropagating its gradients has  $O(NP^{-1}+\log P)$ time complexity.

\subsection{Attenuation Coefficient Estimation}

The attenuation coefficient function $a_c(z)$ is well approximated as a double exponential function~\cite{Akkaynak2019}, characterized using the parameters $v_c, w_c, x_c, y_c \in \mathbb{R}_{\ge0}$ as
\begin{equation}\label{eq:sea-thru-atten-coeff}
    a_c(z) = w_c \exp\left(-v_c \cdot z\right) + y_c \exp\left(-x_c \cdot z\right).
\end{equation}

\begin{figure}
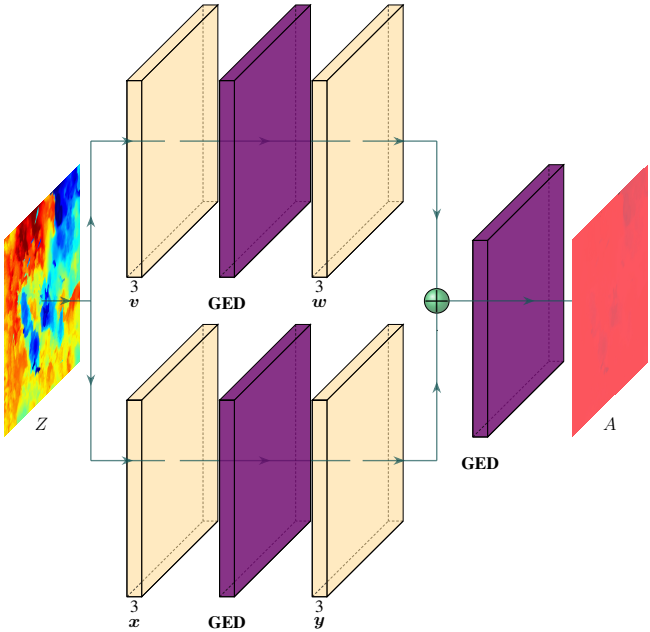

    \centering
    \include{figs/attenuation_net}
    \caption{The attenuation net produces the attenuation map $A$ from a range map $Z$. The kernel parameters in each convolutional layer map to the parameters of the attenuation coefficient function $a(z)$ in Eq.~\eqref{eq:sea-thru-atten-coeff}.}
    \label{fig:attenuation-net}
\end{figure}

\subsubsection{Inference}

From Eqs.~\eqref{eq:sea-thru-attenuation} and~\eqref{eq:sea-thru-atten-coeff}, it follows that
\begin{equation}
    A_c(Z) = \text{GED}\Bigl(Z \cdot \bigl(w_c \cdot \text{GED}(v_c Z) + y_c \cdot \text{GED}(x_c Z)\bigr)\Bigr).
\end{equation}
Accordingly, the attenuation coefficient network depicted in Fig.~\ref{fig:attenuation-net} takes the same structure as the backscatter net, except the CGED and sigmoid activation functions are replaced with GEDs.
% Given the same model parameters, Sea-Thru and DeepSeeColor produce identical attenuation coefficients.
The true-color image $J$ is then estimated as,
\begin{equation}
    \hat{J}_c(i,j) = \hat{D}_c(i,j) \cdot \exp\left(a_c(z_{i,j}) \cdot z_{i,j}\right).
\end{equation}

\subsubsection{Training}

DeepSeeColor trains the attentuation network using a novel composite loss function designed to learn attenuation coefficient parameters similar to those of the Sea-Thru method,
\begin{equation}
    \mathcal{L}_{ac}(\hat{J}; \hat{D}) = \mathcal{L}_\text{saturation}(\hat{J}) + \mathcal{L}_\text{intensity}(\hat{J}) +
    \mathcal{L}_\text{var}(\hat{J}; \hat{D}).
\end{equation}

The saturation loss penalizes over-saturating pixels in the output image, and is defined as
\begin{equation*}
    \mathcal{L}_\text{saturation}(\hat{J}) = \frac{1}{3N} \sum_c \Vert \max\{\hat{J}_c - 1, 0\} \Vert^2,
\end{equation*}
where $N$ is the number of pixels in $\hat{J}$. The intensity loss penalizes image channels which have a very low or very high average intensity, and is defined as
\begin{equation*}
    \mathcal{L}_\text{intensity}(\hat{J}) = \frac{1}{3} \sum_c \left(\frac{1}{N} \sum_{i,j} \left(\hat{J}_c(i,j) - 0.5\right)\right)^2.
\end{equation*}
Finally, the variation loss penalizes changing the variation in each color channel across the image, compared to the variation measured in the direct signal,
\begin{equation*}
    \mathcal{L}_\text{var}(\hat{J}; \hat{D}) = \frac{1}{3} \sum_c \left(s(\hat{J}_c) - s(\hat{D}_c)\right)^2,
\end{equation*}
where $s(M)$ is the standard deviation of the values in $M$.

Minimizing the intensity loss function is similar to following the popular ``gray-world'' approach to white balancing~\cite{Buchsbaum1980}, except the network is constrained to modifying the attenuation coefficients. This drives it to modify each channel's coefficients in a way that stretches contrast but biases the change towards targets far from the camera, and were thus most attenuated. The saturation and variation losses help to avoid very large attenuation coefficients; this is most relevant for images where attenuation has nearly eliminated a color channel such that very large coefficients would be required to recover it. Large coefficients cause over-saturation, and amplify noise in the input image; including the saturation and variation losses mitigates this issue.

This training process has $O(NP^{-1} + \log P)$ time complexity per training iteration when parallelized on a processor with $P\le N$ cores. Furthermore, each iteration produces a color-corrected output image with increased quality, so the method is anytime optimal. For comparison, the Sea-Thru algorithm finds the attenuation model parameters using the LSAC algorithm~\cite{Ebner2013} followed sequentially by a nonlinear optimization algorithm, which are each iterative methods with $O(N)$ time complexity per training iteration. 

\section{Experimental Results}

The DeepSeeColor method has been implemented in PyTorch~\cite{Paszke2019} and demonstrates strong performance on the Sea-Thru dataset~\cite{Akkaynak2019} and on stereo-camera imagery collected with AUV deployments in the US Virgin Islands.

\subsection{Sea-Thru Dataset}

\begin{figure}
    \centering
    \includegraphics[width=\columnwidth]{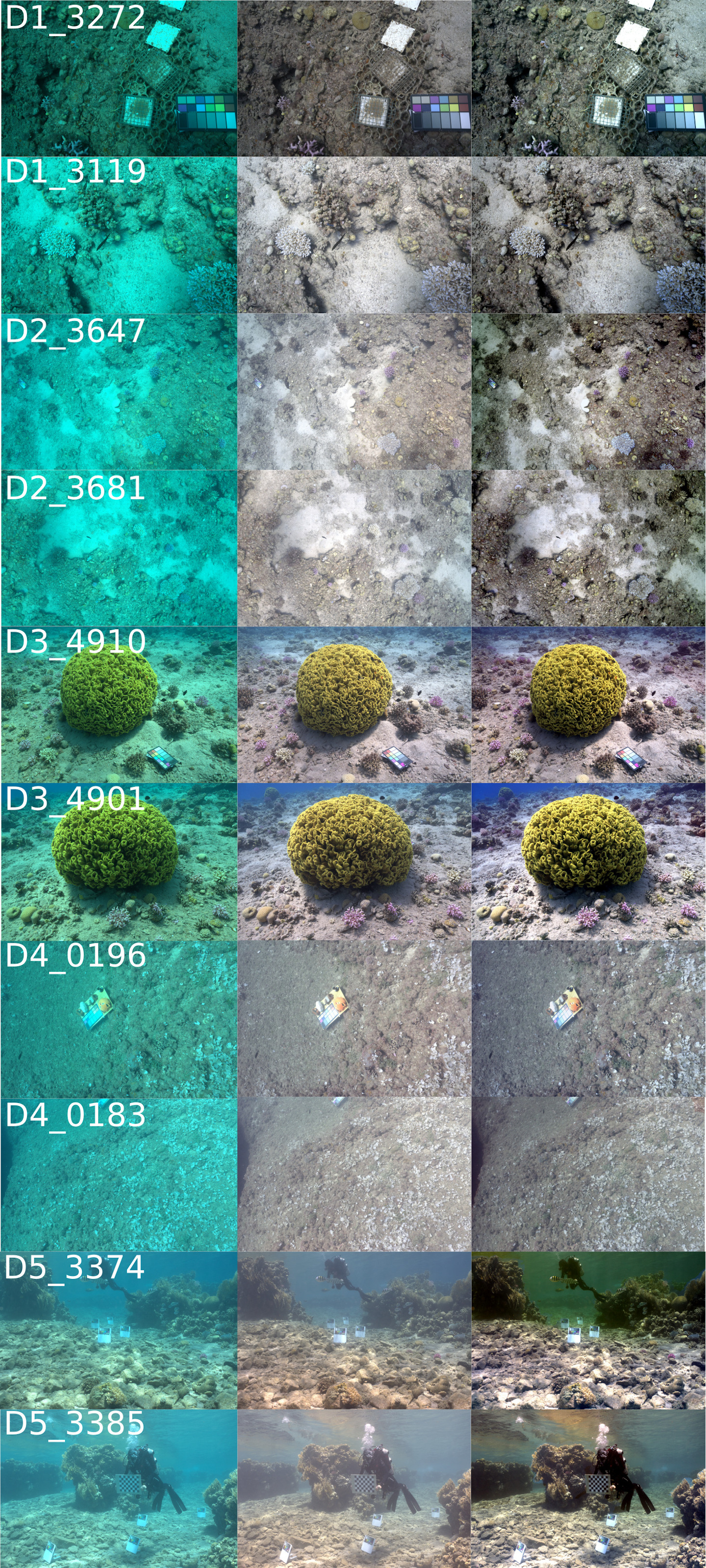}
    \caption{Best viewed on a screen, in color. The left column shows 10 raw images from the Sea-Thru dataset~\cite{Akkaynak2019}. The center column shows the corresponding images after a contrast stretch, while the right column shows the outputs of DeepSeeColor. DeepSeeColor significantly outperforms contrast stretching, especially in scenes with distant targets (e.g. 3647, 3681, and 3385), and achieves comparable performance to Sea-Thru (refer to Fig. 6 in~\cite{Akkaynak2019} for comparison).}
    \label{fig:example-results}
\end{figure}

The Sea-Thru dataset consists of 1157 raw images, each with a corresponding range map generated using structure-from-motion techniques~\cite{Akkaynak2019}. DeepSeeColor was run on all of the images available on the hosting site,\footnote{\url{http://csms.haifa.ac.il/profiles/tTreibitz/datasets/sea_thru/index.html}} and a sampling of the outputs are presented in Fig.~\ref{fig:example-results}.

As in~\cite{Akkaynak2019}, the accuracy of the color reconstruction is estimated using the angular error between true gray and the grayscale patches on the color charts present in many of the dataset images. This angular error is computed at each grayscale patch in the color chart as
\begin{equation}
    \psi(i,j) = \cos^{-1}\left(\frac{\sum_c J_c(i,j)}{\left(3\cdot\sum_c \left[J_c\left(i,j\right)\right]^2\right)^{\frac{1}{2}}}\right).
\end{equation}
The average angular error averaged over the six grayscale patches on each color chart for each of Sea-Thru and DeepSeeColor is presented in Table~\ref{tab:seathru}. As there is no public release of Sea-Thru, its errors are taken directly from~\cite{Akkaynak2019}.

\begin{table}
    \centering
    \caption{Grayscale Patch Mean Angular Error, in degrees.}\label{tab:seathru}
    \begin{tabular}{cccc}
    \toprule 
    Image & Raw & Sea-Thru & DeepSeeColor (ours)\tabularnewline
    \midrule
    \midrule 
    D1\_3272 & 26 & 8 & 14\tabularnewline
    \midrule 
    D2\_3647 & 26 & 8 & 10\tabularnewline
    \midrule 
    D3\_4910 & 22 & 8 & 5\tabularnewline
    \midrule 
    D4\_0209 & 23 & 4 & 4\tabularnewline
    \midrule 
    D5\_3374 & 17/16/15/17 & 4/3/5/3 & 9/10/10/11\tabularnewline
    \bottomrule
    \end{tabular}
\end{table}

\subsection{US Virgin Islands Dataset}

The DeepSeeColor method was also evaluated to process imagery collected in the US Virgin Islands by the CUREE AUV~\cite{Girdhar2023}. The AUV was equipped with a downwards facing color stereo-camera, which enabled the collection of color imagery and range maps like those seen in Figs.~\ref{fig:back_atten} and~\ref{fig:model}. The post-processed imagery has comparable quality to that generated by Sea-Thru, but is generated at a very high rate.

As seen in Table~\ref{tab:deepseathru-runtime}, DeepSeeColor can perform 60 training iterations of each network per second. This corresponds to performing up to 60 training iterations on a single image, or correcting up to 60 images per second once the image formation model parameters have stabilized.
% \footnote{For comparison, the Sea-Thru implementation in footnote~\ref{ft:sea-thru} requires 68s to process a 0.7 megapixel image, and 230s for a 2.4MP image.}
This offers significant flexibility for an AUV to dedicate more processing power to training the DeepSeeColor model when imaging conditions change and the image formation parameters need updating, and less when imaging conditions are stable. Interestingly, the training time is largely insensitive to the input image resolution up to 2.4 megapixels; this is a result of the fixed number of unknown model parameters and of the GPU processing capabilities of modern deep learning frameworks like PyTorch~\cite{Paszke2019}, which process many sections of an image in parallel.

\begin{table}
    \centering
    \caption{DeepSeeColor Runtime.}\label{tab:deepseathru-runtime}
\begin{tabular}{ccccc}
\toprule 
\multicolumn{1}{c}{\# Pixels} & \multicolumn{1}{c}{\begin{tabular}[c]{@{}c@{}}Backscatter\\ (per iter)\end{tabular}} & \multicolumn{1}{c}{\begin{tabular}[c]{@{}c@{}}Attenuation \\ (per iter)\end{tabular}} & \multicolumn{1}{c}{\begin{tabular}[c]{@{}c@{}}Total\\ (per iter)\end{tabular}} & \multicolumn{1}{c}{Max Frequency} \\

% \# Pixels & Backscatter (per iter) & Attenuation (per iter) & Total (per iter) & Max Frequency\tabularnewline
\midrule
\midrule 
0.7M & 6.2 ms & 9.9 ms & 16.1 ms & 62.1 Hz\tabularnewline
\midrule 
2.4M & 6.4 ms & 10.5 ms & 16.9 ms & 59.2 Hz\tabularnewline
\bottomrule
\end{tabular}
\end{table}

% \begin{table*}
%     \centering
%     \caption{DeepSeeColor Runtime.}\label{tab:deepseathru-runtime}
% \begin{tabular}{ccccc}
% \toprule 
% \# Pixels & Backscatter (per iter) & Attenuation (per iter) & Total (per iter) & Max Frequency\tabularnewline
% \midrule
% \midrule 
% 0.7M & 6.2 ms & 9.9 ms & 16.1 ms & 62.1 Hz\tabularnewline
% \midrule 
% 2.4M & 6.4 ms & 10.5 ms & 16.9 ms & 59.2 Hz\tabularnewline
% \bottomrule
% \end{tabular}
% \end{table*}

% TODO (future): show that optimization procedure results in physically valid parameter values? i.e. in the ranges [0,5], etc

\section{Conclusions \& Future Work}

In this paper we have presented DeepSeeColor, a novel color correction method that uses convolutional neural network training operations to learn the parameters of a physics-based underwater image formation model~\cite{Akkaynak2018}. This approach to color correction is more robust than heuristic methods, and thus more appropriate for usage on AUVs performing safety-critical tasks.  We demonstrated our proposed method on the Sea-Thru dataset~\cite{Akkaynak2019}, as well as on images collected during field experiments in the US Virgin Islands.  DeepSeeColor was able to achieve comparable performance in color reconstruction to the Sea-Thru algorithm~\cite{Akkaynak2019}, with significant improvements made in terms of computational complexity. The results support that DeepSeeColor is well-suited for use in realtime preprocessing of imagery collected by an AUV to enable more robust execution of more sophisticated autonomous tasks. In the future, we plan to deploy DeepSeeColor on an AUV to evaluate how access to robust and realtime color correction improves mission performance on various tasks such as place recognition and target tracking.

\section*{Acknowledgements}

The authors thank Nathan McGuire, Seth McCammon, and Levi Cai for their assistance in the development and deployment of the CUREE platform~\cite{Girdhar2023}, which generated much of the data used to develop and test DeepSeeColor.

\clearpage 
\bibliographystyle{IEEEtran}
\bibliography{stewart}

% \newpage
% \listoftodos[Notes]

\end{document}

%% file: tikz_setup.tex
\usepackage{import}
\subimport{layers/}{init}

\usetikzlibrary{positioning}
\usetikzlibrary{3d} %for including external image 
\def\ConvColor{rgb:yellow,5;red,2.5;white,5}

\def\SoftmaxColor{rgb:magenta,5;black,7}   
\def\SumColor{rgb:blue,5;green,15}

%% file: layers/init.tex
%\ProvidesPackage{init}
\usetikzlibrary{quotes,arrows.meta}
\usetikzlibrary{positioning}

\def\edgecolor{rgb:blue,4;red,1;green,4;black,3}
\newcommand{\midarrow}{\tikz \draw[-Stealth,line width =0.8mm,draw=\edgecolor] (-0.3,0) -- ++(0.3,0);}

\usepackage{Ball}
\usepackage{Box}
\usepackage{RightBandedBox}

%% file: figs/backscatter_net.tex
\resizebox{\columnwidth}{0.35\textheight}{%
\begin{tikzpicture}
\tikzstyle{connection}=[ultra thick,every node/.style={sloped,allow upside down},draw=\edgecolor,opacity=0.7]
\tikzstyle{copyconnection}=[ultra thick,every node/.style={sloped,allow upside down},draw={rgb:blue,4;red,1;green,1;black,3},opacity=0.7]
\tikzstyle{every node}=[font=\huge]

\node[canvas is zy plane at x=0] (depth) at (-5,0,0) {\includegraphics[width=8cm,height=8cm]{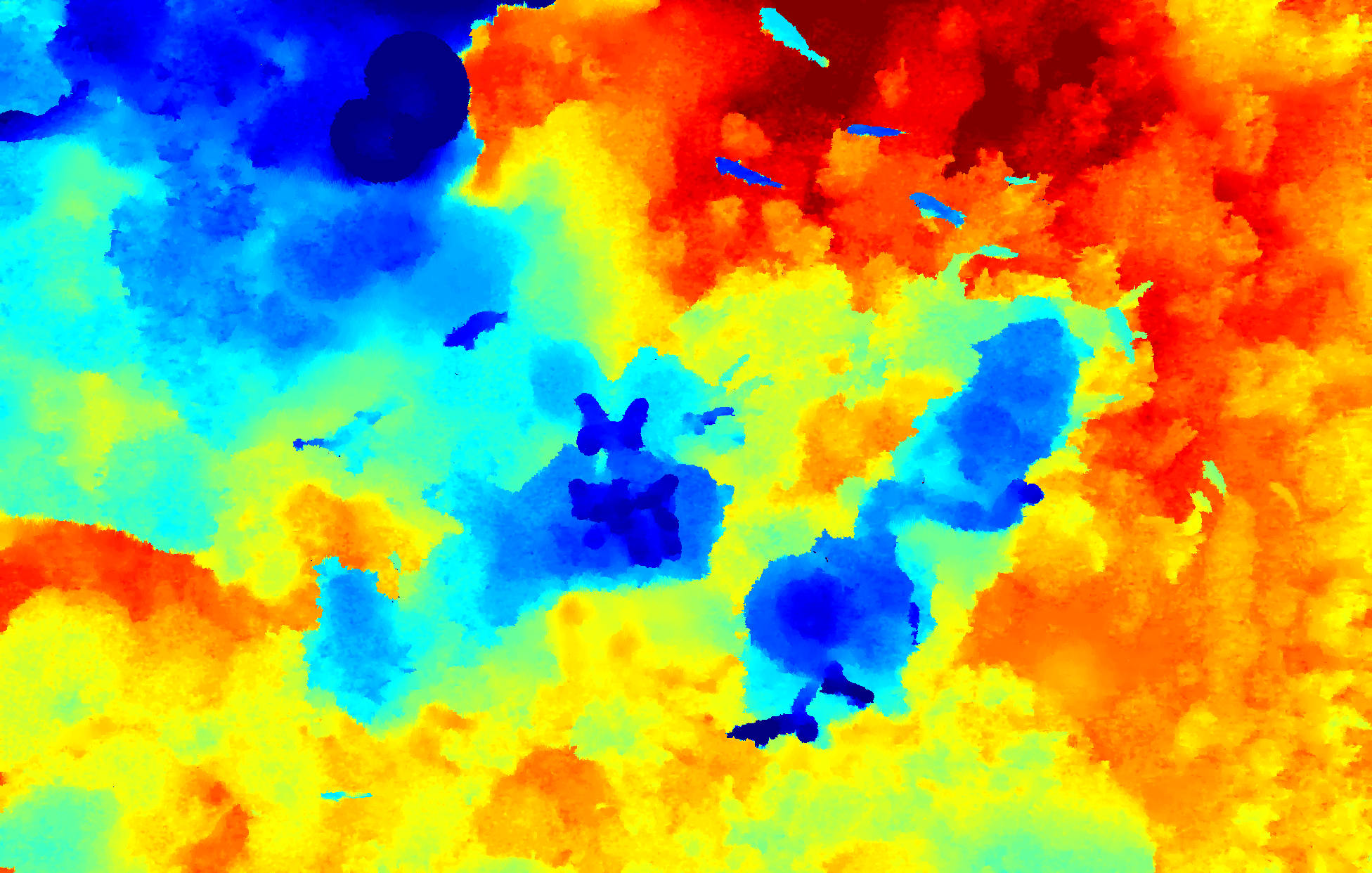}};

\node[] at (-5,-5,0) {$Z$};    

\pic[shift={(0,6.5,0)}] at (0,0,0) 
    {Box={
        name=backscatter,
        caption=$\boldsymbol{\beta}$,
        xlabel={{3, }},
        zlabel=,
        fill=\ConvColor,
        height=40,
        width=3,
        depth=40
        }
    };

\pic[shift={(0, -6.5, 0)}] at (0,0,0) 
    {Box={
        name=residual,
        caption=$\boldsymbol{\alpha}$,
        xlabel={{3, }},
        zlabel=,
        fill=\ConvColor,
        height=40,
        width=3,
        depth=40
        }
    };

\pic[shift={(7.5,6.5,0)}] at (0,0,0) 
    {Box={
        name=Binf,
        caption=$\boldsymbol{\gamma}^\infty$,
        xlabel={{3, }},
        zlabel=,
        fill=\ConvColor,
        height=40,
        width=3,
        depth=40
        }
    };

\pic[shift={(7.5,-6.5,0)}] at (0,0,0) 
    {Box={
        name=Jprime,
        caption=$\boldsymbol{\eta}$,
        xlabel={{3, }},
        zlabel=,
        fill=\ConvColor,
        height=40,
        width=3,
        depth=40
        }
    };

\draw [connection]  (-3, 6.5, 0)    -- node {\midarrow} (0, 6.5, 0);

\draw [connection]  (-3, -6.5, 0)    -- node {\midarrow} (0, -6.5, 0);

\draw [connection]  (-5, 0, 0)    -- node {\midarrow} (-3, 0, 0);

\draw [connection]  (-3, 0, 0)    -- node {\midarrow} (-3, -6.5, 0);

\draw [connection]  (-3, 0, 0)    -- node {\midarrow} (-3, 6.5, 0);

\draw [connection]  (backscatter-east)    -- node {\midarrow} (Binf-west);

\draw [connection]  (residual-east)    -- node {\midarrow} (Jprime-west);

\pic[shift={(11,0,0)}] at (0,0,0) 
    {Ball={
        name=mysum,
        fill=\SumColor,
        caption=,
        opacity=0.6,
        radius=2.5,
        logo=$+$
        }
    };

\draw [connection]  (8.0,6.5,0)    -- node {\midarrow} (11,6.5,0);

\draw [connection]  (8.0,-6.5,0)    -- node {\midarrow} (11,-6.5,0);

\draw [connection]  (11,6.5,0)    -- node {\midarrow} (11,0.5,0);

\draw [connection]  (11,-6.5,0)    -- node {\midarrow} (11,-0.5,0);

\pic[shift={(14, 0, 0)}] at (0,0,0) 
    {Box={
        name=sigmoid,
        caption=Sigmoid,
        xlabel={{" ","dummy"}},
        zlabel=,
        fill=\SoftmaxColor,
        opacity=0.8,
        height=40,
        width=3,
        depth=40
        }
    };

\pic[shift={(3.75, 6.5, 0)}] at (0,0,0) 
    {Box={
        name=top_act,
        caption=CGED,
        xlabel={{" ","dummy"}},
        zlabel=,
        fill=\SoftmaxColor,
        opacity=0.8,
        height=40,
        width=3,
        depth=40
        }
    };

\pic[shift={(3.75, -6.5, 0)}] at (0,0,0) 
    {Box={
        name=bottom_act,
        caption=GED,
        xlabel={{" ","dummy"}},
        zlabel=,
        fill=\SoftmaxColor,
        opacity=0.8,
        height=40,
        width=3,
        depth=40
        }
    };

\draw [connection]  (11.5,0,0)    -- node {\midarrow} (18, 0, 0);

\node[canvas is zy plane at x=0] (output) at (18,0,0) {\includegraphics[width=8cm,height=8cm]{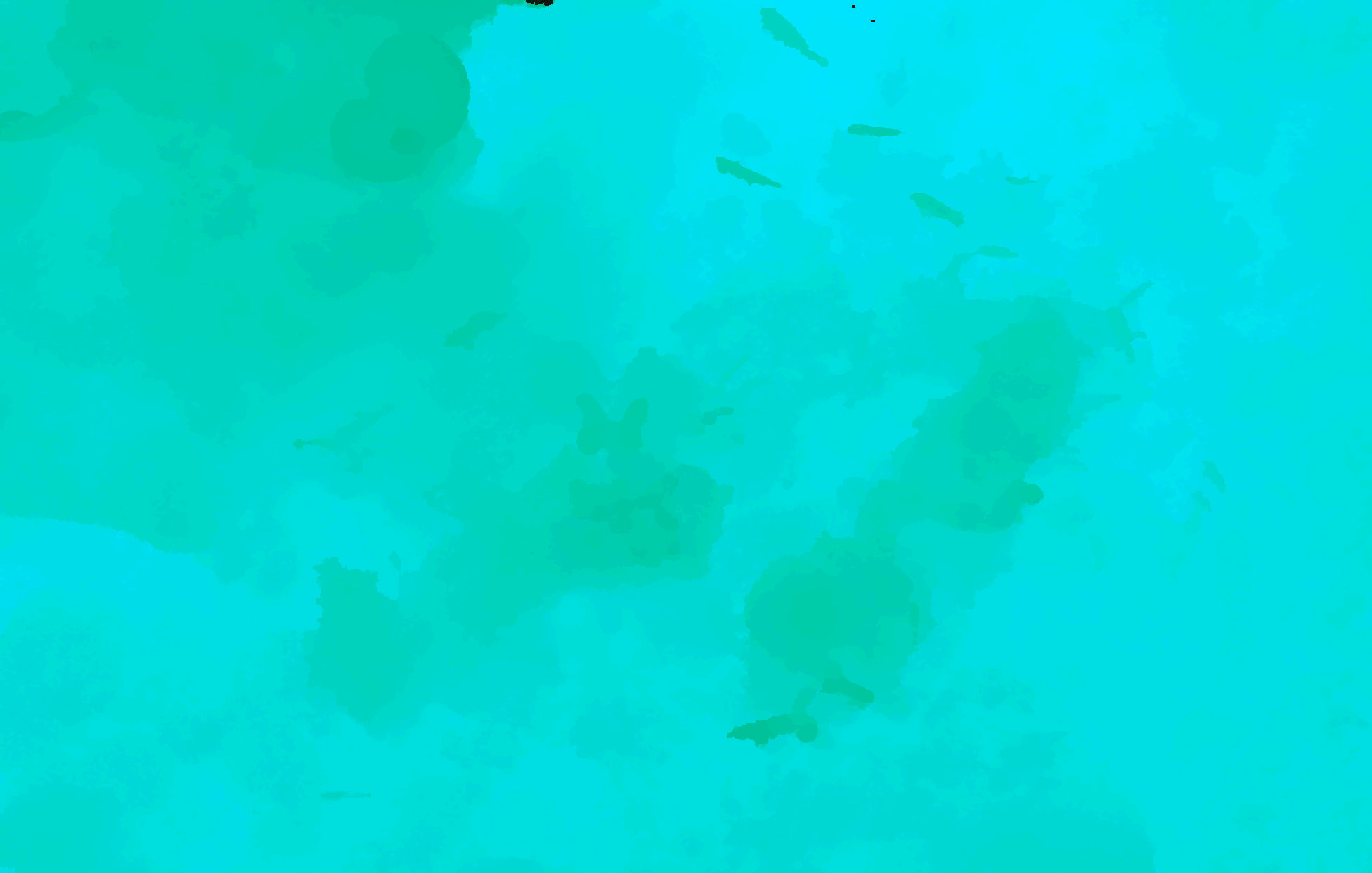}};

\node[] at (18,-5,0) {$\hat{B}$};    

\end{tikzpicture}
}

%% file: figs/attenuation_net.tex
\resizebox{\columnwidth}{0.35\textheight}{%
\begin{tikzpicture}
\tikzstyle{connection}=[ultra thick,every node/.style={sloped,allow upside down},draw=\edgecolor,opacity=0.7]
\tikzstyle{copyconnection}=[ultra thick,every node/.style={sloped,allow upside down},draw={rgb:blue,4;red,1;green,1;black,3},opacity=0.7]
\tikzstyle{every node}=[font=\huge]

\node[canvas is zy plane at x=0] (depth) at (-5,0,0) {\includegraphics[width=8cm,height=8cm]{figs/1659471354-0395682455-warpauv_2.cameras.mapping_cam.left_raw.depth.qoi.jpg}};

\node[] at (-5,-5,0) {$Z$};    

\pic[shift={(0,6.5,0)}] at (0,0,0) 
    {Box={
        name=backscatter,
        caption=$\boldsymbol{v}$,
        xlabel={{3, }},
        zlabel=,
        fill=\ConvColor,
        height=40,
        width=3,
        depth=40
        }
    };

\pic[shift={(0, -6.5, 0)}] at (0,0,0) 
    {Box={
        name=residual,
        caption=$\boldsymbol{x}$,
        xlabel={{3, }},
        zlabel=,
        fill=\ConvColor,
        height=40,
        width=3,
        depth=40
        }
    };

\pic[shift={(7.5,6.5,0)}] at (0,0,0) 
    {Box={
        name=Binf,
        caption=$\boldsymbol{w}$,
        xlabel={{3, }},
        zlabel=,
        fill=\ConvColor,
        height=40,
        width=3,
        depth=40
        }
    };

\pic[shift={(7.5,-6.5,0)}] at (0,0,0) 
    {Box={
        name=Jprime,
        caption=$\boldsymbol{y}$,
        xlabel={{3, }},
        zlabel=,
        fill=\ConvColor,
        height=40,
        width=3,
        depth=40
        }
    };

\draw [connection]  (-3, 6.5, 0)    -- node {\midarrow} (0, 6.5, 0);

\draw [connection]  (-3, -6.5, 0)    -- node {\midarrow} (0, -6.5, 0);

\draw [connection]  (-5, 0, 0)    -- node {\midarrow} (-3, 0, 0);

\draw [connection]  (-3, 0, 0)    -- node {\midarrow} (-3, -6.5, 0);

\draw [connection]  (-3, 0, 0)    -- node {\midarrow} (-3, 6.5, 0);

\draw [connection]  (backscatter-east)    -- node {\midarrow} (Binf-west);

\draw [connection]  (residual-east)    -- node {\midarrow} (Jprime-west);

\pic[shift={(11,0,0)}] at (0,0,0) 
    {Ball={
        name=sum,
        caption=,
        fill=\SumColor,
        opacity=0.6,
        radius=2.5,
        logo=$+$
        }
    };

\draw [connection]  (8.0,6.5,0)    -- node {\midarrow} (11,6.5,0);

\draw [connection]  (8.0,-6.5,0)    -- node {\midarrow} (11,-6.5,0);

\draw [connection]  (11,6.5,0)    -- node {\midarrow} (11,0.5,0);

\draw [connection]  (11,-6.5,0)    -- node {\midarrow} (11,-0.5,0);

\pic[shift={(14, 0, 0)}] at (0,0,0) 
    {Box={
        name=sigmoid,
        caption=GED,
        xlabel={{" ","dummy"}},
        zlabel=,
        fill=\SoftmaxColor,
        opacity=0.8,
        height=40,
        width=3,
        depth=40
        }
    };

\pic[shift={(3.75, 6.5, 0)}] at (0,0,0) 
    {Box={
        name=top_act,
        caption=GED,
        xlabel={{" ","dummy"}},
        zlabel=,
        fill=\SoftmaxColor,
        opacity=0.8,
        height=40,
        width=3,
        depth=40
        }
    };

\pic[shift={(3.75, -6.5, 0)}] at (0,0,0) 
    {Box={
        name=bottom_act,
        caption=GED,
        xlabel={{" ","dummy"}},
        zlabel=,
        fill=\SoftmaxColor,
        opacity=0.8,
        height=40,
        width=3,
        depth=40
        }
    };

\draw [connection]  (11.5,0,0)    -- node {\midarrow} (18, 0, 0);

\node[canvas is zy plane at x=0] (output) at (18,0,0) {\includegraphics[width=8cm,height=8cm]{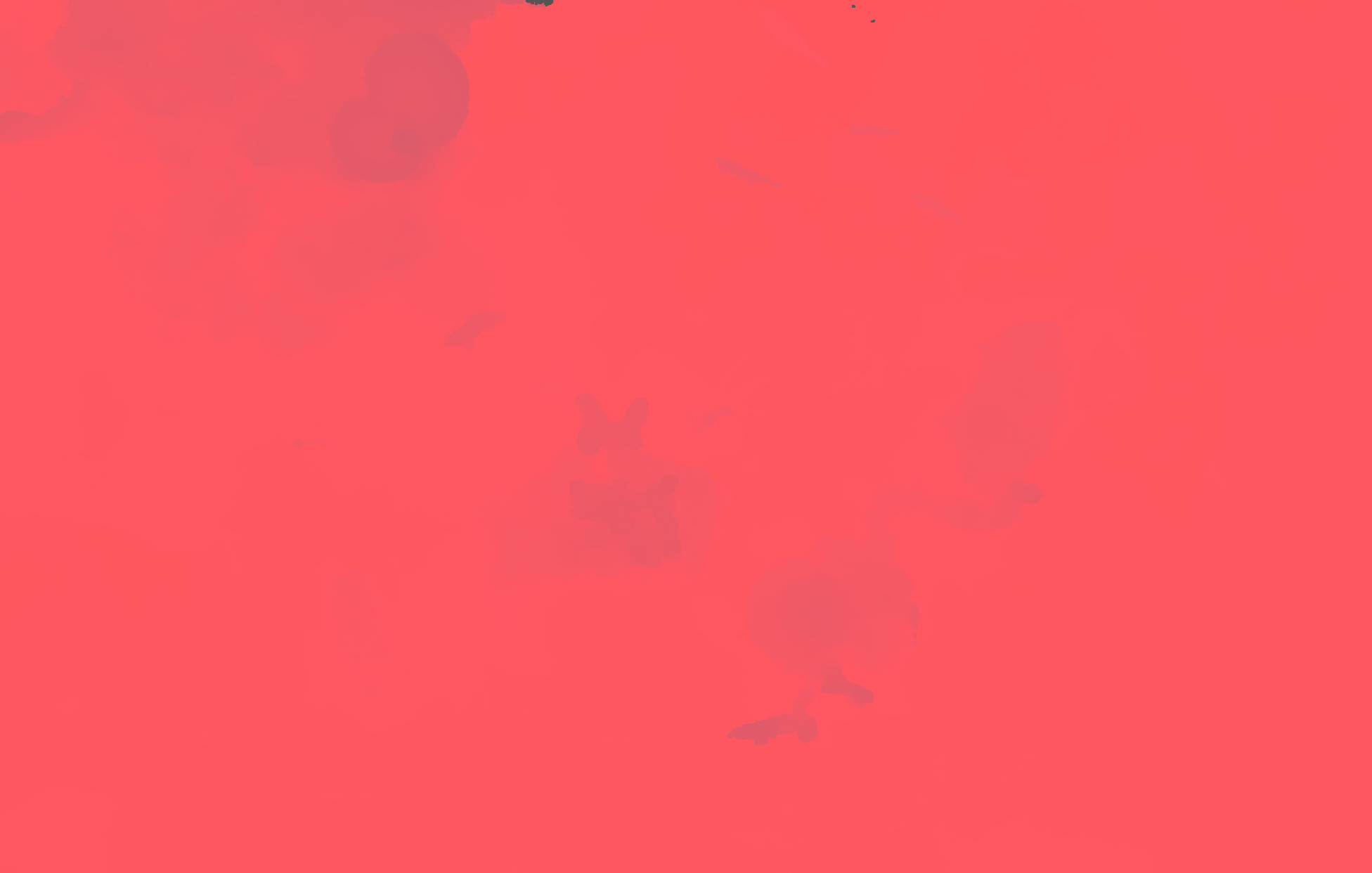}};

\node[] at (18,-5,0) {$A$};    

\end{tikzpicture}
}